\pdfoutput=1

\documentclass[11pt]{article}

\usepackage{acl}

\usepackage{times}
\usepackage{latexsym}

\usepackage[T1]{fontenc}

\usepackage[utf8]{inputenc}

\usepackage{microtype}

\usepackage{inconsolata}

\usepackage{color,xcolor}

\usepackage{pifont}
\usepackage{bm}
\usepackage{amsfonts}

\usepackage{graphicx}
\usepackage[normalem]{ulem}
\usepackage{multirow}
\usepackage{amsmath}
\usepackage{url}

\usepackage{tikz}
\usepackage{xcolor}
\usepackage{tcolorbox}
\usepackage{color,xcolor}

\usepackage{amssymb}
\usepackage{amsopn}
\usepackage{graphicx}
\usepackage{multirow}
\usepackage[english]{babel}
\usepackage{bm}
\usepackage{array}
\usepackage{setspace}
\usepackage{todonotes}

\usepackage{xspace}
\usepackage{amsfonts}
\usepackage{array,multirow}
\usepackage{pgfplots}
\usepackage{tikz}
\usepackage{verbatim}
\usepackage{subfig}
\usepackage{arydshln}
\usepackage{booktabs}
\definecolor{ugreen}{rgb}{0,0.5,0}
\definecolor{mygreen}{RGB}{58,127,88}
\definecolor{iyellow}{RGB}{255,250,205}
\definecolor{ipurple}{RGB}{230,230,250}
\definecolor{myred}{RGB}{160,52,52} 
\definecolor{myblue}{RGB}{30,144,255}
\definecolor{myorange}{RGB}{255,127,80}
\definecolor{mypurple}{RGB}{255,20,147}

\title{Code Comparison Tuning for Code Large Language Models}

\author{Yufan Jiang, \ Qiaozhi He, \  Xiaomin Zhuang, \ Zhihua Wu \\
{\tt jiangyufan2018@outlook.com} \\
}

\begin{document}
\maketitle
\begin{abstract}
We present Code Comparison Tuning (CCT), a simple and effective tuning method for code large language models (Code LLMs) to better handle subtle code errors.
Specifically, we integrate the concept of comparison into instruction tuning, both at the token and sequence levels, enabling the model to discern even the slightest deviations in code.
To compare the original code with an erroneous version containing manually added code errors, we use token-level preference loss for detailed token-level comparisons. Additionally, we combine code segments to create a new instruction tuning sample for sequence-level comparisons, enhancing the model's bug-fixing capability.
Experimental results on the HumanEvalFix benchmark show that CCT surpasses instruction tuning in pass@1 scores by up to 4 points across diverse code LLMs, and extensive analysis demonstrates the effectiveness of our method.
\end{abstract}

\section{Introduction}

Fixing bugs with neural models has become popular among programmers for its powerful capabilities. 
The earliest of these approaches typically consist of multiple individual stages, such as the detection stage and generation stage \cite{lutellier2020coconut,allamanis2021self,yasunaga2021break,mashhadi2021applying,bui2022detect},
whereas Code LLMs successfully address the problem with a simple instruction “Fix the bugs in the code” and achieve competitive performance.

Closed-source LLMs like GPT-4 \cite{openai2023gpt4} have already shown promising results in these code-related tasks.
However, due to high API fees and security problems, exploring how to achieve similar performance using open-source Code LLMs has become a highly meaningful research direction that we focus on in this work.
To ensure responsiveness to human requests, open-source Code LLMs usually undergo a two-step process.
First, they are pre-trained on extensive raw code data, enabling them to acquire a foundational understanding of code patterns and structures \cite{nijkamp2022codegen,fried2022incoder,li2023starcoder,roziere2023code,di2023codefuse}. 
Following pre-training, instruction tuning \cite{wei2021finetuned,ouyang2022training} is employed to align Code LLMs with specific code task instructions provided by humans, such as code completion, bug fixing, or code interpretation \cite{luo2023wizardcoder,shen2023pangu,wang2023codet5+}.

\begin{figure}[!t]
\centering
\includegraphics[width=1.0\linewidth]{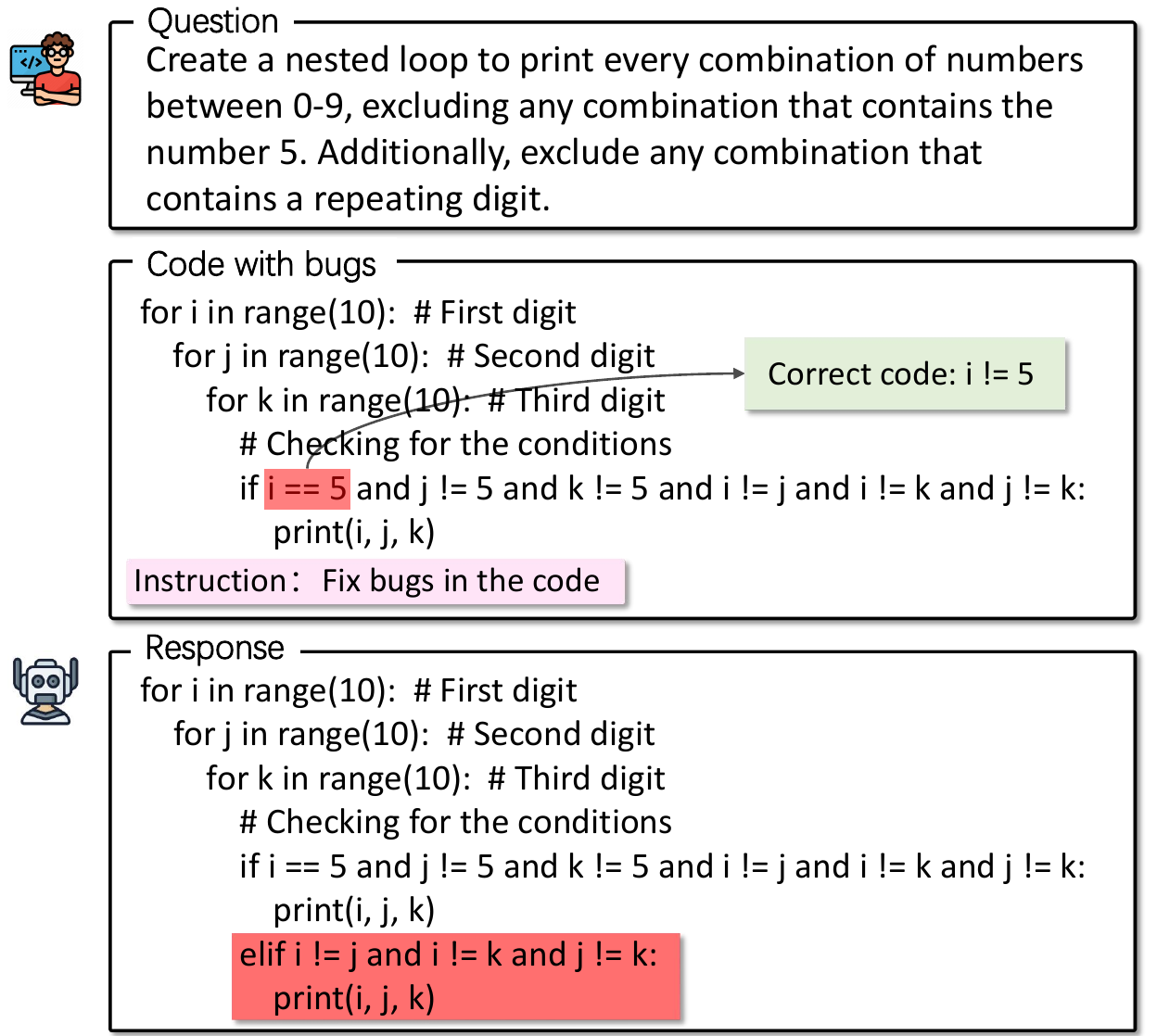}
\caption{
An erroneous bug fix example. Given the code-related issues, users or code language models generate code with bugs. The fine-tuned models tend to introduce additional errors when attempting to fix bugs (red). 
}
\label{fig_example}
\end{figure}

\begin{figure*}[!t]
\centering
\includegraphics[width=1.0\linewidth]{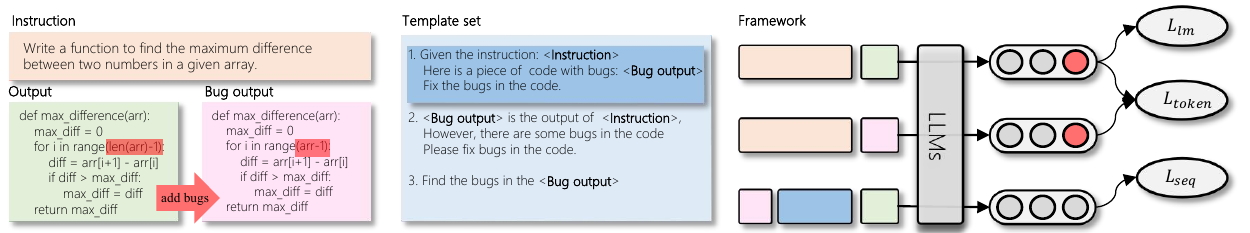}
\caption{
The overall framework of our proposed CCT.
}
\label{fig_model}
\end{figure*}

To further enhance the bug-fixing capabilities of open-source Code LLMs,
some approaches construct specific code-fixing datasets, aiming to bridge the gap between instruction tuning and actual bug fixing \cite{zhang2023self,muennighoff2023octopack}.
Other approaches attempt to integrate code interpreters into the Code LLMs in the form of APIs, enabling real-time code inspection \cite{wang2023leti,bai2023qwen,gou2023critic,chen2023teaching}.
While these solutions have demonstrated effectiveness in practice, teaching Code LLMs to fix bugs remains a challenge. 
Constructing datasets necessitates careful design and collection, making it impractical to cover all error types.
Furthermore, the fine-tuned code models have been proven ineffective in dealing with small changes in the codes \cite{muennighoff2023octopack}. 
When instructed to fix bugs in codes, the models often regenerate the erroneous code or introduce new bugs.
Take the code in Figure \ref{fig_example} as an example. 
Additionally, while code interpreters can assist in identifying syntactic errors, they are unable to detect logical errors within the code.

Here, we present a simple and effective tuning method, namely Code Comparison Tuning (CCT).
This method is specifically designed to heighten the sensitivity of Code LLMs to nuanced variations in code structures. Central to CCT is the integration of a \textit{comparison} mechanism into instruction tuning, realized by creating erroneous versions of each instructive code example. These versions undergo token-level comparative analysis, significantly improving the model's ability to discern and differentiate erroneous code. Additionally, the training dataset is augmented by pairing these generated erroneous codes with their correct forms, as demonstrated in constructs like ``Fixing the error in A results in B'', which further enhances the model's capability of fixing bugs.
Experiments and analysis conducted on the HumanEvalFix benchmark well validate the effectiveness of CCT. Specifically, we observe a substantial improvement over 4 points in pass@1 scores compared to standard instruction tuning on different backbones.

\section{Method}
To make code generative models more sensitive to the errors in the code,
we incorporate two levels of code comparison (token-level and sequence-level) into the instruction tuning of code pre-trained models.
We first give a brief introduction to instruction tuning. 
Then, we introduce two kinds of code comparisons in detail.
While we conducted research on Python in this paper, our approach can be applied to any programming language.

\subsection{Background: Instruction Tuning}
The goal of instruction tuning is to improve the capability of language models in effectively processing instructions expressed in natural languages. 
In general, each instance of instruction-following data begins with "instructions" denoted as $c$, which describes a task, accompanied by a corresponding output $y$ that represents the answer to the given instruction.
The “input” $x$, is the optional context or input for the task. Given the instruction data, the language models are optimized by minimizing the negative log-likelihood of the output $y$:
\begin{equation}
   \mathcal{L}_{lm} = -\frac{1}{\left|y\right|} \sum^{\left|y\right|}_{i} {\rm log}p(y_i|c,x),
   \label{dis1}
\end{equation}

\subsection{Code Comparison Tuning}
We propose two kinds of code comparisons from different perspectives to improve the model's ability to handle error codes.
Specifically, for the code block $t$ in the output $y$, we obtain its counterpart $t'$ by introducing code errors manually.
Then, we perform comparisons between $t$ and $t'$ at both token level and sequence level.

To construct code containing bugs, we initially extract code blocks from the output $y$. Subsequently, we introduce bugs by either randomly replacing or deleting elements such as variables, functions, and operators within these code segments. Bug examples are in Appendix \ref{sec:appendix}.
With examples of the correct code and error code, the model is optimized to locate the bugs and fix them.

\paragraph{Token-level Comparison}
Previous studies usually provide supervision signals to code language models by training samples in the format of bug fixes, in order to guide models on how to repair bugs.
However, this type of sequence-based training sample causes the model to ignore more granular-level differences between code snippets, which results in the degeneration of the model's ability to repair errors.
To tackle this problem, we adopt a token-level comparison loss \citep{zeng2023tim}
to teach models to be aware of the changes in each token. 

Formally, given code $t$ and its counterpart $t'$, the token-level comparison loss is defined as:
\begin{align}
    \nonumber \mathcal{L}_{token} = -\frac{1}{M - I} \sum^{N}_{i=I} \max(&0, -r_{\theta}(h^{t}_i) + \\ 
    &r_{\theta}(h^{t'}_i) + 1.0),
\end{align}
where $I$ represents the index starting from the first differing segment between sequences $t$ and $t'$, and $M$ is the maximum length of two sequences. 
The hidden state of each token $i$ is denoted as $h_i$ and we add a linear head $r_{\theta}$ that converts the hidden state to a scalar.

\paragraph{Sequence-level Comparison}
Beyond mastering token-level distinctions, our approach integrates both $t$ and $t'$ within a single sentence, facilitating the model's acquisition of sequence-level repair skills.
Specifically, we first create a set of templates designed to transform comparative code pairs into coherent instructional data.
Then we convert the code pairs to instruction-tuning style by randomly sampling a template from $T$. 
All the templates are illustrated in Appendix \ref{sec:appendix2}.
Finally, the sequence-level comparison example is used to optimize the language model via Eq.(\ref{dis1}) with the associated loss denoted as $\mathcal{L}_{seq}$.

\subsection{Overall Training Objective}
The overall training objective is defined as:
\begin{equation}
   \mathcal{L} = \mathcal{L}_{lm} + \alpha * \mathcal{L}_{token} + \beta * \mathcal{L}_{seq},
\end{equation}
where $\alpha$ and $\beta$ are non-negative hyper-parameters to balance the effect of each loss term. In this paper, we set $\alpha$ and $\beta$ to 2.0 and 0.5, respectively

\section{Experiments}
\subsection{Datasets}
We conducted experiments on Evol-Instruct-Code-80k dataset \footnote{https://github.com/nickrosh/evol-teacher} licensed by Apache-2.0.
The dataset is created following the process described in the WizardCoder Paper \cite{luo2023wizardcoder}.
We extracted data from code written in Python to use as our instruction data.
\begin{table}[!t]
\centering
\small
\begin{spacing}{0.95}
\begin{tabular}{lcc}
\toprule
{\bf Model} & Params & Pass@1 \\
\midrule
{\it Closed-source LLMs} \\
{\bf ChatGPT} & - & {39.6} \\
{\bf GPT-4} & >=175B & {47.0} \\
\midrule
{\it Open-source LLMs} \\
InstructCodeT5+* & 16B & 2.7 \\
BLOOMZ* & 176B & 16.6 \\
StarCoder* & 15.5B & 8.7 \\
CodeLlama* & 13B & 15.2 \\
CodeGeeX2* & 6B & 15.9 \\
OctoCoder* & 15.5B & 30.4 \\
WizardCoder & 15.5B & 31.8 \\
WizardCoder-Python-13B & 13B & 47.7 \\
\midrule
{\it StarCoder backbone} \\
{\bf Instruct tuning} & 15.5B & 33.7 \\ 
{\bf CCT-StarCoder (Ours)} & 15.5B & 38.3 \\ 
\midrule
{\it CodeLlama-Python-13B backbone} \\
{\bf Instruct tuning} & 13B & 43.5 \\ 
{\bf CCT-CodeLlama (Ours)} & 13B & {\bf 47.7} \\ 
\bottomrule
\end{tabular}
\end{spacing}
\caption{
\label{tab_results}
{\bf Pass@1 (\%) performance on HumanEvalFix.}
Models with * denote that we directly report the scores from the corresponding paper}
\end{table}
To verify the effectiveness of our proposed approach, we evaluated CCT on the HumanEvalFix \cite{muennighoff2023octopack} which is proposed to task models to fix the bugs in function.
It contains 164 HumanEval solutions across all 6 languages (984 total bugs) and the errors are manually inserted into the code.

\subsection{Baselines \& Settings}
We mainly experimented on CodeLlama-13b-Python \cite{roziere2023code} and StarCoder \cite{li2023starcoder} in this work. 
Additionally, we report the results of InstructCodeT5+, BLOOMZ, CodeGeeX2, StarCoder, OctoCoder and WizardCoder-Python-13B \cite{muennighoff2022crosslingual,wang2023codet5+,li2023starcoder,zheng2023codegeex,luo2023wizardcoder}.
We also report the results of closed-source models such as ChatGPT and GPT-4  which can be accessed via API.

To facilitate a fair and consistent evaluation, we fine-tuned all models for 1 epoch with a batch size of 64. The learning rate was set to 2e-5 and the weight decay parameter was set to 0.0.
For evaluation, we used the pass@1 metric \cite{chen2021evaluating}.
Similar to \citet{muennighoff2023octopack}, we used a sampling temperature of 0.2 and
$top_p$ = 0.95 to estimate pass@1. We generated n = 20 samples, which is enough to get reliable pass@1 estimates \cite{li2023starcoder}.

\begin{table}[!t]
\centering
\small
\begin{spacing}{0.95}
\begin{tabular}{lcc}
\toprule
{\bf Model} & Pass@1 \\
\midrule
{\bf Instruct tuning} & {43.53$\pm$0.49} \\
\ \ \ \ w Sequence-level data& {45.76$\pm$0.23} \\
{\bf CCT-CodeLlama} & {47.71$\pm$0.39} \\
\ \ \ \ w/o $\mathcal{L}_{seq}$ & {44.56$\pm$0.61} \\
\ \ \ \ w/o $\mathcal{L}_{token}$ & {45.83$\pm$0.32} \\
\bottomrule
\end{tabular}
\end{spacing}
\caption{
\label{tab_results_abl}
{\bf Ablation study. } We run each experiment 3 times with different random seeds and report mean and standard deviation .}
\end{table}

\subsection{Results}
Table \ref{tab_results} shows the results of several models on HumanEvalFix.
We can see that most open-source code LLMs struggle with handling subtle code changes and instructing tuning can substantially enhance their performance.
Our Code comparison tuning significantly outperforms instruct tuning on both StarCoder and CodeLlama-Python-13B backbone, leading to an average of 4 Pass@1 scores improvement

At the same time, CCT achieves comparable results to its open-source competitors of the same size.
These results demonstrate the effectiveness of the proposed code comparison method.
Although CCT has surpassed GPT4 on HumanEvalFix, we still need to conduct further testing for evaluation.  We leave this issue for future study.

\subsection{Ablation Study}
To analyze the impact of different components of CCT, we investigate the following variants: 1) {\it CCT} {w/o} $\mathcal{L}_{seq}$, removing the sequence-level comparison; 2) {\it CCT} {w/o} $\mathcal{L}_{token}$, removing the token-level comparison;
Additionally, we utilize the data generated from the sequence-level comparison phase to create instruction fine-tuning data and mix it together with original data to fine-tune the model which denotes as {\it w Sequence-level data}.
We take CodeLlama-Python-13B as the backbone.

The results are listed in Table \ref{tab_results_abl}.
The degradation of {\it CCT} {w/o} $\mathcal{L}_{seq}$ and {\it CCT} {w/o} $\mathcal{L}_{token}$ indicate that code LLMs can improve their ability to learn how to fix errors in code by leveraging the code comparison in both token and sequence levels.
While {\it w Sequence-level data} performs better than the standard instruct tuning, there is still room for improvement as our CCT achieved even better results.
This suggests that our proposed method goes beyond just data augmentation, as it incorporates comparison during fine-tuning to enhance the effectiveness and efficiency of code LLMs.
\begin{table}[!t]
\centering
\small
\begin{spacing}{0.95}
\begin{tabular}{lcc}
\toprule
{\bf Model} & Pass@1 \\
\midrule
{\it Closed-source LLMs} \\
GPT-4 & {88.4} \\
\midrule
{\it Open-source LLMs} \\
WizardCoder-Python-13B & 60.37 \\
Instruct tuning & 63.26 \\ 
{\bf CCT-CodeLlama} & {66.1} \\
\bottomrule
\end{tabular}
\end{spacing}
\caption{
\label{tab_results_docs}
{\bf Pass@1 (\%) performance on HumanEvalFixDocs.}}
\end{table}

\subsection{Results on HumanEvalFixDocs}
HumanEvalFixDocs \cite{muennighoff2023octopack} provides docstrings as the source of ground truth for the model to fix the buggy function which is generally easier for models than HumanEvalFix.
From Table \ref{tab_results_docs}, we see that our CCT performs significantly better than instruction fine-tuning and other open-source models. 
However, it also reveals a notable performance gap compared with GPT4, an aspect we aim to explore in our future research.

\subsection{Effect of Corpus Size}
In this experiment, we study the impact of data sizes on CCT by sampling different percentages of the instructing dataset.
Figure \ref{fig_corpus_size} shows the comparison between our CCT and instruct tuning under different data sizes.
We see that, when the amount of data used gradually decreases, our CCT still maintains a strong performance.
Surprisingly, with only 20\% of the data, CCT can achieve a pass@1 score of 43,
demonstrating the data efficiency of our proposed method.
\begin{figure}[!t]
  \centering
  \begin{tikzpicture}[scale = 0.85]
    \footnotesize{
      \begin{axis}[
      ymajorgrids,
  xmajorgrids,
  grid style=dashed,
      width=.50\textwidth,
      height=.20\textwidth,
      legend style={at={(0.20,0.12)}, anchor=south west},
      xlabel={\scriptsize{Data size}},
      ylabel={\scriptsize{Pass@1}},
      ylabel style={yshift=-1em},
      xlabel style={yshift=0.0em},
      ymin=28,ymax=50, ytick={30, 35, 40, 45, 50},
      xmin=18,xmax=100,xtick={20, 40, 60, 80, 100},
      legend style={yshift=2pt, legend plot pos=right, legend columns=3 ,font=\scriptsize,cells={anchor=west}}
      ]

      \addplot[red!70,mark=diamond*,line width=1pt] coordinates {(20,43.8) (40,44.51) (60,46.2) (80,46.4) (100,47.7)};
      \addlegendentry{\scriptsize CCT}
      \addplot[blue!70,mark=diamond*,line width=1pt] coordinates {(20,33) (40,41.5) (60,42.5) (80,43.4) (100,43.5)};
      \addlegendentry{\scriptsize Instruct tuning}
      \end{axis}
     }
  \end{tikzpicture}
  \caption{{\bf Effect of Instruction dataset size.} We report pass@1 under
  different sizes of instructing datasets.}
  \label{fig_corpus_size}
\end{figure}

\section{Conclusions}
In this work, we enhance the ability of code LLMs to fix bugs by integrating code comparison during instruct tuning.
We consider both token-level and sequence-level comparisons to make code models more sensitive to the small changes in the code.
Experiments and analyses validate the effectiveness of our model.
We plan to extend our method to more programming languages and conduct tests on a wider range of test sets in our future study.


\section*{Limitations}
There are still a few drawbacks of our approach that need further investigation.
The construction method we use for generating error code snippets is relatively simple. Introducing more complex construction methods is necessary to provide the model with additional comparative information.
Second, more bug-fixing tests are needed, including a wider range of programming languages and a greater variety of error types.
We leave these investigations for future work.
While we have achieved remarkable results in the evaluation metrics of the code repair task, there is still an ongoing need for continuous research and dedicated efforts to enhance how code-pretrained models can better assist programmers in handling code-related tasks.
\bibliography{anthology,custom}

\appendix

\section{Construction of erroneous codes}
\label{sec:appendix}
\begin{figure}[!t]
\centering
\includegraphics[width=1.0\linewidth]{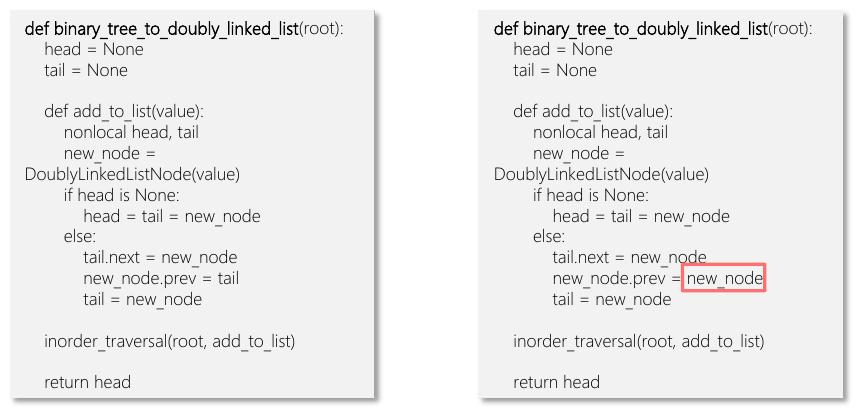}
\caption{Variable misuse bug example. The buggy code (right) incorrectly uses 'newcode'.
}
\label{fig_example4}
\end{figure}

\begin{figure}[!t]
\centering
\includegraphics[width=1.0\linewidth]{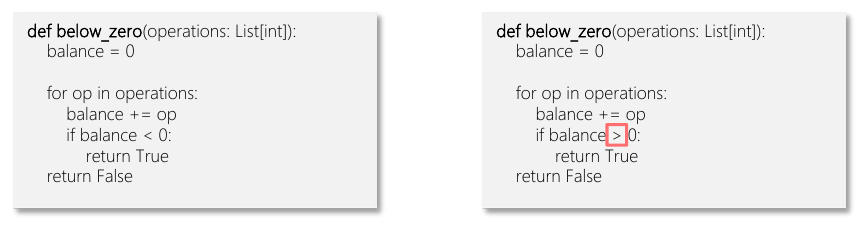}
\caption{Operator misuse bug example. The buggy code (right) incorrectly uses 'greater than'.
}
\label{fig_example5}
\end{figure}

\begin{figure}[!t]
\centering
\includegraphics[width=1.0\linewidth]{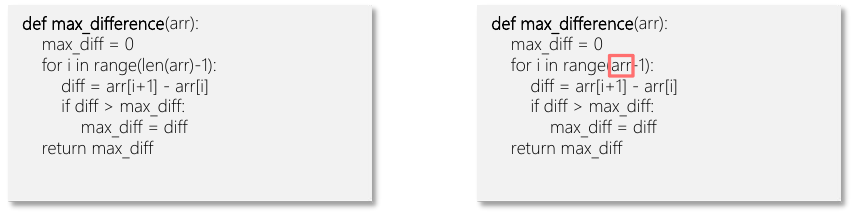}
\caption{Function missing bug example. The buggy code (right) removes 'len()' function.
}
\label{fig_example6}
\end{figure}

In this work, we focus on incorporating token-level bugs into codes.
We add the following types of bugs:
1) Misuse variables in the code, as shown in \ref{fig_example4}.
2) Misuse operators in the code, as shown in \ref{fig_example5}.
3) Miss functions in the code, as shown in \ref{fig_example6}.
More methods can be tried to create erroneous examples, such as using GPT-4 for generation. We will leave this part of the work for the future.

\section{Templates for Sequence-level Comparison}
\label{sec:appendix2}
The templates we used to construct sequence-level comparison examples are illustrated in \ref{tab_template}
\begin{table}[!t]
\centering
\small
\begin{spacing}{1.5}
\begin{tabular}{l}
\toprule
\midrule 
{\bf Templates}  \\
\midrule 
Given the instruction: <Instruction> \\
Here is a piece of code with bugs: <Bug output> \\
Fix the bugs in the code. \\
\midrule 
<Bug output> is the code implementation of <Instruction>, \\
However, there are some bugs in the code \\
Please fix bugs in the code. \\
\midrule 
Find the bugs in the <Bug output> \\
\midrule 
\bottomrule 
\end{tabular}
\end{spacing}
\caption{
\label{tab_template}
{Templates for constructing sequence-level comparison examples}.
}
\end{table}

\end{document}